\documentclass{article}
\usepackage{spconf,amsmath,graphicx}
\usepackage{cite}
\usepackage{amsmath,amssymb,amsfonts}
\usepackage{algorithmic}
\usepackage{graphicx}
\usepackage{textcomp}
\usepackage{xcolor}
\usepackage{booktabs}
\usepackage[table]{xcolor}
\usepackage{hyperref}

\newcommand{\methodname}{RevFFN}

\title{RevFFN: Memory-Efficient Full-Parameter Fine-Tuning of Mixture-of-Experts LLMs with Reversible Blocks}
\name{Ningyuan Liu\qquad Jing Yang\qquad Kaitong Cai\qquad Keze Wang}
\address{Sun Yat-sen University}

\author{Anonymous Submission}

\begin{document}

\maketitle

\begin{abstract}
Full Fine-Tuning is a pivotal technique for adapting Large Language Models (LLMs) to downstream tasks, yet it incurs substantial memory overhead due to the necessity of caching extensive intermediate activations for backpropagation. This bottleneck renders the full fine-tuning of contemporary, large-scale LLMs exceedingly challenging. While existing distributed training frameworks like DeepSpeed mitigate this issue through techniques such as ZeRO and FSDP, which leverage multi-GPU memory or CPU offloading, these solutions often come at the cost of increased hardware requirements and diminished training speed. To address this core challenge, we introduce \methodname, a novel memory-efficient fine-tuning paradigm. \methodname{} employs meticulously designed reversible Transformer blocks, enabling the reconstruction of input activations for any given layer from its output during backpropagation, thereby obviating the need to store the majority of intermediate activations in memory. This approach, while preserving the integrity of the Mixture-of-Experts (MoE) architecture, drastically reduces the peak memory required for full fine-tuning. Consequently, it facilitates efficient full fine-tuning on a single consumer or server-grade GPU.
\end{abstract}

\section{Introduction}

In recent years, Large Language Models (LLMs) such as GPT-4\cite{openai2024gpt4technicalreport}, LLaMA\cite{touvron2023llamaopenefficientfoundation}, and Qwen\cite{z1} have achieved revolutionary advancements in Natural Language Processing and multimodal domains. The inherent scalability of the Transformer architecture\cite{z2,z3} implies that, under similar designs, larger models typically exhibit superior generalization and reasoning capabilities. However, when adapting these models to specific downstream tasks via full fine-tuning, conventional training methods encounter a severe memory bottleneck. The root of this problem lies in the backpropagation-based optimization algorithms central to modern deep learning frameworks. For instance, the widely-used Adam optimizer\cite{z4,Z5,z6} requires caching the activations from every layer during the forward pass to compute gradients. The size of these activations scales proportionally with model parameter count and batch size, resulting in a prohibitive memory footprint for LLMs with billions or even hundreds of billions of parameters.
To surmount this challenge, researchers have proposed various memory optimization techniques. Memory sharding and offloading, exemplified by DeepSpeed's ZeRO\cite{DBLP:journals/corr/abs-2004-08818} and PyTorch's FSDP, effectively reduce the VRAM pressure on a single GPU by distributing the model across multiple GPUs or offloading them to host DRAM. However, such methods do not decrease the total memory required; instead, they introduce significant inter-device data communication, demanding high bus bandwidth and leading to an reduction in training speed. An alternative approach is Parameter-Efficient Fine-Tuning (PEFT), such as LoRA, which drastically cuts memory usage by training only a small set of adapter parameters, but at the expense of the potential performance gains from a full parameter update. This brings a core challenge to the forefront: can we design a novel Transformer module that, at the cost of a modest increase in computation, fundamentally eliminates the need to store most activations, thereby revolutionizing memory efficiency?

To solve this challenge, we introduce \methodname, a Transformer block architected on the principles of reversible networks. \methodname{} partitions the hidden state of a standard Transformer layer into two streams and processes them through a unique coupled update rule. This design guarantees that the input to the block can be precisely derived from its output, allowing for the dynamic recomputation of activations during the backward pass without prior caching. We implement our method on the Qwen1.5-MoE model, designing learnable projection layers to adapt pre-trained weights while fully preserving the MoE structure. Experiments show that \methodname{} enables full-parameter fine-tuning of LLMs on a single GPU with lower memory overhead and better performance than PEFT methods.
\begin{figure*}[h!]
    \includegraphics[width=\textwidth]{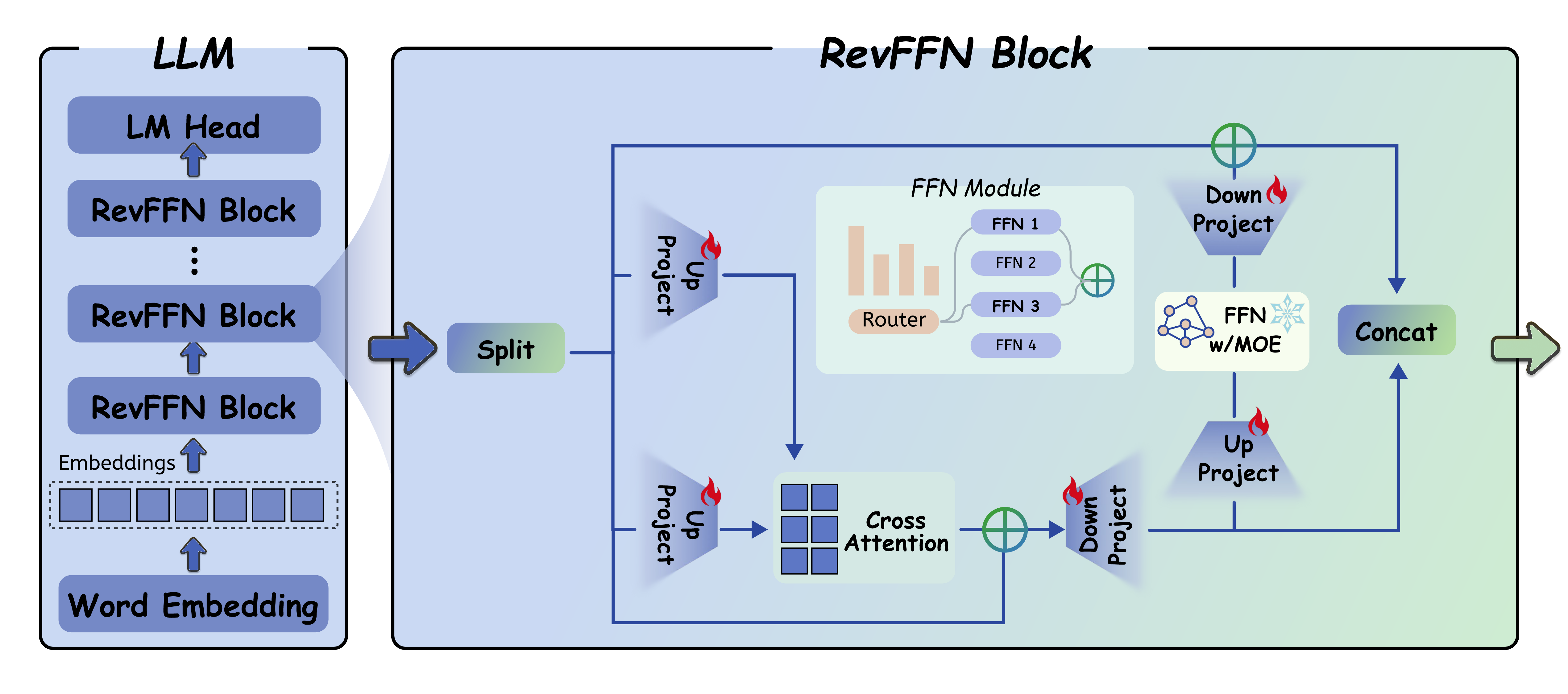}
    
\caption{\methodname{} architecture: hidden states split, processed by Cross-Attention and MoE, projected, and concatenated as output.}    
    \label{fig:revffn_architecture}
\end{figure*}

\section{Related Work}

\subsection{Memory-Efficient Training.}
LoRA\cite{DBLP:journals/corr/abs-2106-09685} and its variants (e.g., QLoRA\cite{dettmers2023qloraefficientfinetuningquantized}) exemplify parameter-efficient fine-tuning (PEFT), achieving low memory usage by inserting low-rank adapters and training only these modules.  
While efficient, PEFT methods update only a small fraction of parameters, which can limit model capacity on complex tasks.  
Our goal is to preserve the flexibility of full fine-tuning while maintaining PEFT-level memory efficiency.

DeepSpeed ZeRO and Fully Sharded Data Parallel (FSDP) address memory bottlenecks by partitioning model states across devices.  
In its most advanced stage, ZeRO-3 ensures that each GPU stores only a fraction of the parameters.  
Activation checkpointing further reduces memory via recomputation, but these techniques usually require multi-GPU clusters and are often constrained by interconnect bandwidth when offloading activations.

\subsection{Reversible Networks.}
Reversible networks, first introduced by RevNet, draw inspiration from Ordinary Differential Equation (ODE) solvers\cite{z8,z10,z11}.  
Their central idea is to design layers whose inputs can be exactly reconstructed from outputs.  
A standard RevNet block partitions the input into two parts, each updated using a function of the other, enabling straightforward inversion.  
This eliminates the need to store intermediate activations, as they can be recomputed during backpropagation.  
Although extensively studied in computer vision, applying reversibility to large-scale Transformer-based Mixture-of-Experts (MoE) models\cite{z12} for full fine-tuning remains both novel and challenging.  
Our \methodname{} leverages this principle, specifically adapted to Transformer and MoE architectures.

\section{Methodology}
\label{sec:method}

\vspace{-4mm}

\methodname{} combines the memory–efficiency of \emph{reversible networks} with the computation–efficiency of \emph{mixture-of-experts (MoE) large language models}.  
Our implementation starts from the public \texttt{Qwen2Moe} checkpoints in Hugging Face Transformers\cite{bai2023qwentechnicalreport}, then wraps each decoder layer with a reversible scaffold and a pair of lightweight projection adapters.

\vspace{-4mm}

\subsection{\methodname{} Reversible Block}\label{sec:rev_block}
A classical Transformer decoder layer consists of
a Self-Attention sub-layer;  
a position-wise Feed-Forward Network (FFN) or MoE block;  
LayerNorm + residual connections around both.
Reversible networks rearrange these components so that the input of every layer can be exactly reconstructed from its output, eliminating the need to cache hidden states during back-propagation and therefore cutting peak memory roughly in half.
\noindent\textbf{Forward pass.} %
Let \(H\in\mathbb{R}^{B\times S\times d_{\mathrm{model}}}\) be the current hidden tensor (\(B\)=batch, \(S\)=sequence length).
We split it along the feature dimension,
\[
H = \bigl[X_1,\;X_2\bigr],\qquad
X_1,X_2 \in \mathbb{R}^{B\times S\times \tfrac{1}{2}d_{\text{model}}},
\]
and update the two halves in a \emph{coupled} fashion:
\begin{align}
Y_1 &= X_1 + \underbrace{\mathrm{Attn}\!\bigl(
      \mathrm{Norm}(X_1),
      \mathrm{Norm}(X_2),
      \mathrm{Norm}(X_2)\bigr)}_{\text{cross-branch attention}},
      \label{eq:fwd_y1}\\[3pt]
Y_2 &= X_2 + \underbrace{\mathrm{MLP}\!\bigl(
      \mathrm{Norm}(Y_1)\bigr)}_{\text{FFN or MoE}}.
      \label{eq:fwd_y2}
\end{align}
Finally we concatenate $H_{\text{out}} = [Y_1,Y_2]$ and pass it to the next layer.
This layout differs from the standard residual stack in two ways:
1. Attention queries come from the \emph{left} stream ($X_1$) while keys/values come from the \emph{right} stream ($X_2$);  
2. The FFN/MoE is driven solely by the updated left stream ($Y_1$).
The asymmetry encourages information exchange between halves and was found to stabilise training compared with a symmetric variant\cite{DBLP:journals/corr/abs-2001-04451}.
\noindent\textbf{Inverse pass and memory saving.} %
Because Eqs.~\eqref{eq:fwd_y1}--\eqref{eq:fwd_y2} form a bijection, activations can be recomputed on demand:
\[
\begin{aligned}
\hat{X}_2 &= Y_2 - \mathrm{MLP}\!\bigl(\mathrm{Norm}(Y_1)\bigr),\\
\hat{X}_1 &= Y_1 - \mathrm{Attn}\!\bigl(
              \mathrm{Norm}(\hat{X}_1),
              \mathrm{Norm}(\hat{X}_2),
              \mathrm{Norm}(\hat{X}_2)\bigr).
\end{aligned}
\]
The second line contains $\hat{X}_1$ on both sides; in practice we run one fixed-point iteration starting from $\mathrm{Norm}(Y_1)$.  
This yields a reconstruction error below machine epsilon while incurring negligible extra compute.
\subsection{Projection Adapters for Pre-trained Modules}\label{sec:proj}

A central challenge arises because the pre-trained attention and MLP weights expect $d_{\mathrm{model}}$-dimensional inputs, 
while each reversible stream provides only $d_{\mathrm{model}}/2$ features.  
Retraining these large matrices from scratch would negate the benefit of leveraging pre-training.  
To bridge this mismatch, we introduce lightweight projection adapters: two small linear layers placed before and after each pre-trained block,
\[
P_{\uparrow}: \mathbb{R}^{d/2}\!\to\!\mathbb{R}^{d}, 
\qquad
P_{\downarrow}: \mathbb{R}^{d}\!\to\!\mathbb{R}^{d/2}.
\]

\vspace{-4mm}

\[
\small
\begin{aligned}
\mathrm{Attn}(X_1,X_2)
  &= P_{\downarrow}\!\Bigl(
     \mathrm{Attn}_{\text{pt}}\bigl(
       \underbrace{P_{\uparrow}(X_1)}_{\text{query }Q},
       \underbrace{P_{\uparrow}(X_2)}_{\text{key }K},
       \underbrace{P_{\uparrow}(X_2)}_{\text{value }V}\bigr)\Bigr), \\[-2pt]
\mathrm{MLP}(Y_1)
  &= P_{\downarrow}\!\Bigl(
     \mathrm{MLP}_{\text{pt}}\bigl(
       \underbrace{P_{\uparrow}(Y_1)}_{\text{projection}}\bigr)\Bigr).
\end{aligned}
\]

\noindent\textbf{Advantages.}  
(1) All computationally intensive operations (attention matrix multiplications, MoE routing, expert MLPs) still occur in the original high-dimensional space, thus preserving model capacity.  
(2) The additional cost is minimal: the two projections introduce only $O(d^2)$ parameters, which is negligible compared to the backbone.  
(3) The design is \emph{plug-and-play}: any off-the-shelf Transformer layer can be wrapped with adapters without modifying its internals.
\subsection{Two-Stage Training Schedule}\label{sec:training}
We employ a curriculum that first aligns sub-spaces with adapters, then fine-tunes the full model.
\textbf{Stage 1: Adapter warm-up.}  
Freeze all pre-trained weights (including MoE experts and routers) and train only $P_{\uparrow},P_{\downarrow}$ for a few epochs with a small learning rate.  
This quickly adapts the reversible streams to the backbone manifold while avoiding catastrophic forgetting.
\textbf{Stage 2: Joint fine-tuning.}  
Unfreeze the Transformer layers and continue end-to-end training.  
MoE gating networks remain frozen to preserve routing stability\cite{fedus2022switchtransformersscalingtrillion}; only expert weights and adapters are updated.  
This schedule yields smoother loss curves and higher downstream accuracy than training everything at once.
Overall, \methodname{} attains large-scale capacity, strong downstream performance, and memory savings.

\section{Experiment}
To validate the efficacy and efficiency of \methodname{}, we designed a comprehensive set of experiments to evaluate its performance against several established and state-of-the-art fine-tuning paradigms. We focus on memory consumption, training speed, and performance on downstream benchmarks.

\subsection{Experimental Setup}
\textbf{Base Model:} We use ``Qwen/Qwen1.5-MoE-A2.7B'' from Hugging Face as our pre-trained base model \cite{bai2023qwentechnicalreport}. This model features a Mixture-of-Experts architecture with 2.7 billion activated parameters.

\noindent \textbf{Hardware \& Software:} All experiments were conducted on a single \textbf{NVIDIA H800} GPU with 80GB of VRAM, using PyTorch and the Transformers library.

\noindent \textbf{Dataset:} We use the ``databricks-dolly-15k'' dataset, a popular open-source instruction-following dataset, for all fine-tuning tasks.

\noindent \textbf{Baselines for Comparison:}
We compare our method against a comprehensive set of baselines.

\textit{Parameter-Efficient Fine-Tuning (PEFT) Methods:}
\textbf{LoRA:} A widely-used method that trains low-rank adapters, serving as a low-memory, high-throughput baseline \cite{DBLP:journals/corr/abs-2106-09685}.
\textbf{DoRA:} An enhancement over LoRA that decomposes pre-trained weights into magnitude and direction for more effective fine-tuning \cite{liu2024doraweightdecomposedlowrankadaptation}.
\textbf{(IA)$^3$:} A highly efficient PEFT method that introduces learned vectors to rescale activations, modifying the model's behavior with minimal new parameters \cite{liu2022fewshotparameterefficientfinetuningbetter}.

\textit{Memory-Efficient Full-Parameter Fine-Tuning Methods:}
\textbf{SFT w/ Activation Checkpointing:} The conventional approach to full fine-tuning, augmented with gradient checkpointing to reduce memory at the cost of recomputation \cite{DBLP:journals/corr/ChenXZG16}.
\textbf{LoMo:} A memory-efficient optimizer that fuses gradient computation and parameter updates, significantly reducing memory usage for optimizer states \cite{lv2024parameterfinetuninglargelanguage}.
\textbf{GaLore:} A state-of-the-art memory-efficient strategy that uses a Gradient Low-Rank Projection optimizer to enable full-parameter tuning with reduced memory overhead \cite{zhao2024galorememoryefficientllmtraining}.

\noindent \textbf{Evaluation Metrics:}
\textbf{Peak VRAM Usage:} The maximum GPU memory allocated during training, measured in Gigabytes (GB). This is the primary metric for memory efficiency.
 \textbf{Training Throughput:} The number of samples processed per second (samples/sec), indicating training speed.
\textbf{Downstream Task Performance:} We evaluate the models on a diverse set of benchmarks to assess their reasoning, multilingual, and conversational abilities. For reasoning and knowledge, we use \textbf{MMLU}\cite{DBLP:journals/corr/abs-2009-03300} to measure multitask language understanding and \textbf{GSM8K} \cite{gomez2017reversible} to test multi-step mathematical reasoning. To evaluate cross-lingual capabilities, we report the average performance on a \textbf{Multilingual} benchmark suite. Finally, we assess conversational ability using \textbf{MT-Bench} \cite{zheng2023judging}, which provides a score for multi-turn instruction-following and chat performance.

\subsection{Main Results}
We fine-tuned the Qwen1.5-MoE-A2.7B model using each method on the Dolly dataset for a fixed number of steps. The batch size for each method was maximized to fit within the 80GB VRAM constraint. The efficiency and performance results are summarized in Table \ref{tab:memory_speed} and Table \ref{tab:performance}, respectively.

\vspace{-5mm}

\begin{table}[ht!]
    \centering
    \caption{Memory and Speed Comparison on a Single H800 GPU. Lower VRAM is better; higher throughput is better.}
    \label{tab:memory_speed}
    \scalebox{0.65}{
    \begin{tabular}{l c c}
        \toprule
        \textbf{Method} & \textbf{Peak VRAM (GB)} $\downarrow$ & \textbf{Throughput (samples/s)} $\uparrow$ \\
        \midrule
        \textit{PEFT Methods} \\
        LoRA \cite{DBLP:journals/corr/abs-2106-09685} & 18.2 & \textbf{75.4} \\
        DoRA \cite{liu2024doraweightdecomposedlowrankadaptation} & 19.5 & 71.8 \\
        (IA)$^3$ \cite{liu2022fewshotparameterefficientfinetuningbetter} & 17.9 & 74.1 \\
        \midrule
        \textit{Full-Parameter Fine-Tuning Methods} \\
        SFT + Checkpointing & 65.4 & 19.7 \\
        LOMO \cite{lv2024parameterfinetuninglargelanguage} & 42.2 & 17.3 \\
        GaLore \cite{zhao2024galorememoryefficientllmtraining} & 45.1 & 35.2 \\
        \rowcolor{gray!25}
        \textbf{\methodname{}} & \textbf{39.5} & 24.6 \\
        \bottomrule
    \end{tabular}
    }
\end{table}

\vspace{-8mm}

\subsection{Analysis of Results}
The results demonstrate that \methodname{} achieves a superior trade-off between memory efficiency and model performance.

\textbf{Memory and Speed Efficiency}
As shown in Table \ref{tab:memory_speed}, PEFT methods like LoRA and DoRA exhibit the lowest memory footprint and highest throughput. However, our proposed \methodname{} achieves a remarkable \textbf{49\% reduction in peak VRAM usage} compared to SFT with Checkpointing. It is also more memory-efficient than GaLore. The memory saving is the core contribution of our reversible architecture. In terms of speed, the recomputation inherent in \methodname{} introduces an overhead, resulting in lower throughput than other methods. However, its throughput is still faster than SFT, as our reversible design is more computationally efficient.

\begin{table}[ht!]
    \centering
    \caption{Downstream Benchmark Performance. Higher is better for all metrics.}
    \label{tab:performance}
    \scalebox{0.58}{
    \begin{tabular}{l c c c c}
        \toprule
        \textbf{Method} & \textbf{MMLU (\%)} $\uparrow$ & \textbf{GSM8K (\%)} $\uparrow$ & \textbf{Multilingual (\%)} $\uparrow$ & \textbf{MT-Bench (score)} $\uparrow$\\
        \midrule
        Base Model & 62.4 & 61.2 & 40.4 & 6.25 \\
        \midrule
        LoRA & 65.2 & 71.5 & 38.5 & 7.18\\
        DoRA & 65.7 & 70.8 & 38.9 & 7.25 \\
        (IA)$^3$ & 65.0 & 70.2 & 38.2 & 7.15 \\
        \midrule
        SFT + Checkpointing & 66.1 & 74.8 & \textbf{39.5} & 7.52\\
        LOMO & 66.2 & 74.6 & 39.3 & 7.50 \\
        GaLore & 66.3 & 74.2 & 39.2 & 7.46 \\
        \rowcolor{gray!25}
        \textbf{\methodname{}} & \textbf{66.7} & \textbf{75.1} & 38.8 & \textbf{7.65} \\
        \bottomrule
    \end{tabular}}
\end{table}

\textbf{Downstream Task Performance}
Table \ref{tab:performance} highlights the performance benefits of full-parameter updates. \methodname{} consistently outperforms all other methods across the MMLU, GSM8K, Multilingual and MT-Bench benchmarks. Notably, it surpasses even the strong SFT baseline, suggesting that our reversible architecture does not compromise, and may even slightly enhance, model expressiveness. The significant performance gap between \methodname{} and leading PEFT methods like DoRA validates the importance of updating all model parameters to unlock its full capabilities on complex downstream tasks

\subsection{Ablation Study}

\vspace{-4mm}
\begin{table}[h!]
    \centering
    \caption{Ablation study of the two-stage training strategy on the MMLU benchmark. Higher is better.}
    \label{tab:ablation}
    \begin{tabular}{l c}
        \toprule
        \textbf{Configuration} & \textbf{MMLU (\%)} $\uparrow$ \\
        \midrule
        \textbf{\methodname{} (Full Method)}  & \textbf{66.7} \\
        w/o Stage 1 (Joint Training)  & 57.1 \\
        w/o Stage 2 (Projections Only) & 54.5 \\
        \bottomrule
    \end{tabular}
\end{table}

To understand the contribution of each component in our proposed method, we conducted an ablation study focusing on our two-stage training strategy. The results, evaluated on the MMLU benchmark, are presented in Table \ref{tab:ablation}.

The results show that each stage is crucial for optimal performance. Skipping the initial warm-up phase (Stage 1) by training all parameters jointly from the start leads to a noticeable performance drop, suggesting this phase is vital for preventing training instability. Furthermore, omitting full-parameter fine-tuning (Stage 2) and only training the projection layers—a configuration analogous to an adapter-based PEFT method—results in a significant degradation. 

These results highlight a clear division of labor between the two stages: Stage 1 stabilizes the representation space, while Stage 2 fully unleashes model capacity. Without either stage, the system either fails to converge effectively or is overly constrained in expressiveness. This ablation validates our two-stage strategy as an essential component of the \methodname{} framework, ensuring both stability and performance gains over naive training alternatives.

\section{Conclusion}
We proposed RevFFN, an efficient full fine-tuning method for MOE LLMs based on reversible networks. By splitting hidden states into two reversible streams, RevFFN eliminates the need to cache large activations during backpropagation. Experiments show that it reduces memory barriers, enabling single-GPU full fine-tuning with superior performance. Future work includes scaling to larger models and exploring synergies with techniques such as and knowledge distillation.  
Overall, our results highlight that RevFFN offers a practical path to balance memory efficiency with full-parameter adaptability, making single-GPU fine-tuning feasible for increasingly large LLMs.

\bibliographystyle{IEEEbib}
\bibliography{references} 

@article{DBLP:journals/corr/abs-2106-09685,
  author       = {Edward J. Hu and
                  Yelong Shen and
                  Phillip Wallis and
                  Zeyuan Allen{-}Zhu and
                  Yuanzhi Li and
                  Shean Wang and
                  Weizhu Chen},
  title        = {LoRA: Low-Rank Adaptation of Large Language Models},
  journal      = {CoRR},
  volume       = {abs/2106.09685},
  year         = {2021},
  url          = {https://arxiv.org/abs/2106.09685},
  eprinttype    = {arXiv},
  eprint       = {2106.09685},
  bibsource    = {dblp computer science bibliography, https://dblp.org}
}

@misc{liu2024doraweightdecomposedlowrankadaptation,
      title={DoRA: Weight-Decomposed Low-Rank Adaptation}, 
      author={Shih-Yang Liu and Chien-Yi Wang and Hongxu Yin and Pavlo Molchanov and Yu-Chiang Frank Wang and Kwang-Ting Cheng and Min-Hung Chen},
      year={2024},
      eprint={2402.09353},
      archivePrefix={arXiv},
      primaryClass={cs.CL},
      url={https://arxiv.org/abs/2402.09353}, 
}

@misc{liu2022fewshotparameterefficientfinetuningbetter,
      title={Few-Shot Parameter-Efficient Fine-Tuning is Better and Cheaper than In-Context Learning}, 
      author={Haokun Liu and Derek Tam and Mohammed Muqeeth and Jay Mohta and Tenghao Huang and Mohit Bansal and Colin Raffel},
      year={2022},
      eprint={2205.05638},
      archivePrefix={arXiv},
      primaryClass={cs.LG},
      url={https://arxiv.org/abs/2205.05638}, 
}

@article{DBLP:journals/corr/ChenXZG16,
  author       = {Tianqi Chen and
                  Bing Xu and
                  Chiyuan Zhang and
                  Carlos Guestrin},
  title        = {Training Deep Nets with Sublinear Memory Cost},
  journal      = {CoRR},
  volume       = {abs/1604.06174},
  year         = {2016},
  url          = {http://arxiv.org/abs/1604.06174},
  eprinttype    = {arXiv},
  eprint       = {1604.06174},
  bibsource    = {dblp computer science bibliography, https://dblp.org}
}

@misc{lv2024parameterfinetuninglargelanguage,
      title={Full Parameter Fine-tuning for Large Language Models with Limited Resources}, 
      author={Kai Lv and Yuqing Yang and Tengxiao Liu and Qinghui Gao and Qipeng Guo and Xipeng Qiu},
      year={2024},
      eprint={2306.09782},
      archivePrefix={arXiv},
      primaryClass={cs.CL},
      url={https://arxiv.org/abs/2306.09782}, 
}

@misc{zhao2024galorememoryefficientllmtraining,
      title={GaLore: Memory-Efficient LLM Training by Gradient Low-Rank Projection}, 
      author={Jiawei Zhao and Zhenyu Zhang and Beidi Chen},
      year={2024},
      eprint={2403.03507},
      archivePrefix={arXiv},
      primaryClass={cs.LG},
      url={https://arxiv.org/abs/2403.03507}, 
}

@article{DBLP:journals/corr/abs-2009-03300,
  author       = {Dan Hendrycks and
                  Collin Burns and
                  Steven Basart},
  title        = {Measuring Massive Multitask Language Understanding},
  journal      = {CoRR},
  volume       = {abs/2009.03300},
  year         = {2020},
  url          = {https://arxiv.org/abs/2009.03300},
  eprinttype    = {arXiv},
  eprint       = {2009.03300},
  bibsource    = {dblp computer science bibliography, https://dblp.org}
}

@misc{zheng2023judging,
  title  = {Judging LLM-as-a-Judge with MT-Bench and Chatbot Arena},
  author = {Zheng, Lianmin and others},
  year   = {2023},
  note   = {arXiv:2306.05685}
}

@misc{bai2023qwentechnicalreport,
      title={Qwen Technical Report}, 
      author={Jinze Bai and Shuai Bai and Yunfei Chu},
      year={2023},
      eprint={2309.16609},
      archivePrefix={arXiv},
      primaryClass={cs.CL},
      url={https://arxiv.org/abs/2309.16609}, 
}

@misc{touvron2023llamaopenefficientfoundation,
      title={LLaMA: Open and Efficient Foundation Language Models}, 
      author={Hugo Touvron and Thibaut Lavril and Gautier Izacard},
      year={2023},
      eprint={2302.13971},
      archivePrefix={arXiv},
      primaryClass={cs.CL},
      url={https://arxiv.org/abs/2302.13971}, 
}

@article{DBLP:journals/corr/abs-2004-08818,
  author       = {Bart M. P. Jansen and
                  Jari J. H. de Kroon},
  title        = {Preprocessing Vertex-Deletion Problems: Characterizing Graph Properties
                  by Low-Rank Adjacencies},
  journal      = {CoRR},
  volume       = {abs/2004.08818},
  year         = {2020},
  url          = {https://arxiv.org/abs/2004.08818},
  eprinttype    = {arXiv},
  eprint       = {2004.08818},
  bibsource    = {dblp computer science bibliography, https://dblp.org}
}

@misc{dettmers2023qloraefficientfinetuningquantized,
      title={QLoRA: Efficient Finetuning of Quantized LLMs}, 
      author={Tim Dettmers and Artidoro Pagnoni and Ari Holtzman and Luke Zettlemoyer},
      year={2023},
      eprint={2305.14314},
      archivePrefix={arXiv},
      primaryClass={cs.LG},
      url={https://arxiv.org/abs/2305.14314}, 
}

@misc{gomez2017reversible,
  title  = {The Reversible Residual Network: Backpropagation Without Storing Activations},
  author = {Gomez, Aidan N},
  year   = {2017},
  note   = {arXiv:1707.04585}
}

@misc{openai2024gpt4technicalreport,
  title = {GPT-4 Technical Report},
  author = {OpenAI},
  year = {2024},
  eprint = {2303.08774},
  archivePrefix = {arXiv},
  primaryClass = {cs.CL},
  url = {https://arxiv.org/abs/2303.08774},
}

@article{DBLP:journals/corr/abs-2001-04451,
  author       = {Nikita Kitaev and
                  Lukasz Kaiser and
                  Anselm Levskaya},
  title        = {Reformer: The Efficient Transformer},
  journal      = {CoRR},
  volume       = {abs/2001.04451},
  year         = {2020},
  url          = {https://arxiv.org/abs/2001.04451},
  eprinttype    = {arXiv},
  eprint       = {2001.04451},
  bibsource    = {dblp computer science bibliography, https://dblp.org}
}

@misc{fedus2022switchtransformersscalingtrillion,
      title={Switch Transformers: Scaling to Trillion Parameter Models with Simple and Efficient Sparsity}, 
      author={William Fedus and Barret Zoph and Noam Shazeer},
      year={2022},
      eprint={2101.03961},
      archivePrefix={arXiv},
      primaryClass={cs.LG},
      url={https://arxiv.org/abs/2101.03961}, 
}

@inproceedings{
z1,
title={{KABB}: Knowledge-Aware Bayesian Bandits for Dynamic Expert Coordination in Multi-Agent Systems},
author={Jusheng Zhang and Zimeng Huang and Yijia Fan and Ningyuan Liu and Mingyan Li and Zhuojie Yang and Jiawei Yao and Jian Wang and Keze Wang},
booktitle={Forty-second International Conference on Machine Learning},
year={2025},
url={https://openreview.net/forum?id=AKvy9a4jho}
}

@inproceedings{
z2,
title={{GAM}-Agent: Game-Theoretic and Uncertainty-Aware Collaboration for Complex Visual Reasoning},
author={Jusheng Zhang and Yijia Fan and Wenjun Lin and Ruiqi Chen and Haoyi Jiang and Wenhao Chai and Jian Wang and Keze Wang},
booktitle={The Thirty-ninth Annual Conference on Neural Information Processing Systems},
year={2025},
url={https://openreview.net/forum?id=EKJhU5ioSo}
}

@misc{z3,
      title={CF-VLM:CounterFactual Vision-Language Fine-tuning}, 
      author={Jusheng Zhang and Kaitong Cai and Yijia Fan and Jian Wang and Keze Wang},
      year={2025},
      eprint={2506.17267},
      archivePrefix={arXiv},
      primaryClass={cs.LG},
      url={https://arxiv.org/abs/2506.17267}, 
}

@inproceedings{
z4,
title={{MAT}-Agent: Adaptive Multi-Agent Training Optimization},
author={Jusheng Zhang and Kaitong Cai and Yijia Fan and Ningyuan Liu and Keze Wang},
booktitle={The Thirty-ninth Annual Conference on Neural Information Processing Systems},
year={2025},
url={https://openreview.net/forum?id=YDWRTYgR79}
}

@inproceedings{
Z5,
title={Tri-{MARF}: A Tri-Modal Multi-Agent Responsive Framework for Comprehensive 3D Object Annotation},
author={Jusheng Zhang and Yijia Fan and Zimo Wen and Jian Wang and Keze Wang},
booktitle={The Thirty-ninth Annual Conference on Neural Information Processing Systems},
year={2025},
url={https://openreview.net/forum?id=YGIbwfNWot}
}

@misc{z6,
      title={MM-CoT:A Benchmark for Probing Visual Chain-of-Thought Reasoning in Multimodal Models}, 
      author={Jusheng Zhang and Kaitong Cai and Xiaoyang Guo and Sidi Liu and Qinhan Lv and Ruiqi Chen and Jing Yang and Yijia Fan and Xiaofei Sun and Jian Wang and Ziliang Chen and Liang Lin and Keze Wang},
      year={2025},
      eprint={2512.08228},
      archivePrefix={arXiv},
      primaryClass={cs.CV},
      url={https://arxiv.org/abs/2512.08228}, 
}

@misc{z8,
      title={Failure-Driven Workflow Refinement}, 
      author={Jusheng Zhang and Kaitong Cai and Qinglin Zeng and Ningyuan Liu and Stephen Fan and Ziliang Chen and Keze Wang},
      year={2025},
      eprint={2510.10035},
      archivePrefix={arXiv},
      primaryClass={cs.AI},
      url={https://arxiv.org/abs/2510.10035}, 
}

@misc{z10,
      title={Learning Dynamics of VLM Finetuning}, 
      author={Jusheng Zhang and Kaitong Cai and Jing Yang and Keze Wang},
      year={2025},
      eprint={2510.11978},
      archivePrefix={arXiv},
      primaryClass={cs.LG},
      url={https://arxiv.org/abs/2510.11978}, 
}

@misc{z11,
      title={DrDiff: Dynamic Routing Diffusion with Hierarchical Attention for Breaking the Efficiency-Quality Trade-off}, 
      author={Jusheng Zhang and Yijia Fan and Kaitong Cai and Zimeng Huang and Xiaofei Sun and Jian Wang and Chengpei Tang and Keze Wang},
      year={2025},
      eprint={2509.02785},
      archivePrefix={arXiv},
      primaryClass={cs.CL},
      url={https://arxiv.org/abs/2509.02785}, 
}

@misc{z12,
      title={Kolmogorov-Arnold Fourier Networks}, 
      author={Jusheng Zhang and Yijia Fan and Kaitong Cai and Keze Wang},
      year={2025},
      eprint={2502.06018},
      archivePrefix={arXiv},
      primaryClass={cs.LG},
      url={https://arxiv.org/abs/2502.06018}, 
}

\end{document}